\documentclass{article}
\usepackage[OT1]{fontenc}
\usepackage{spconf,amsmath,graphicx,booktabs,hyperref}
\DeclareMathSizes{9.1}{9}{7}{5}
\hypersetup{hidelinks}

\title{CAUSALGLOSS: MULTI-EVIDENCE ACTION CONTROL FOR A SIMULTANEOUS
MANDARIN SPEECH-TO-ASL GLOSS CASCADE}

\name{Mingze Li$^{\star}$ \qquad Satoshi Nakamura$^{\dagger}$}
\address{$^{\star}$School of Data Science, The Chinese University of Hong Kong, Shenzhen, China\\
         $^{\dagger}$School of Artificial Intelligence, The Chinese University of Hong Kong, Shenzhen, China}

\begin{document}
\AddToHookNext{shipout/foreground}{%
  \put(54bp,-742bp){%
    \parbox[t]{505bp}{\normalfont\small
      This work has been submitted to the IEEE for possible publication.
      Copyright may be transferred without notice, after which this version may no longer be accessible.}%
  }%
}
\maketitle
\fontsize{9.8pt}{11.76pt}\selectfont
\emergencystretch=2em

\begingroup
\fontsize{9.8pt}{10pt}\selectfont
\begin{abstract}
Simultaneous Mandarin speech-to-American Sign Language (ASL) gloss translation
requires deciding when an evolving utterance supports committing each gloss.
Fixed Wait-$k$ cannot adapt to source--gloss reordering, while training only on
reference histories causes exposure bias. We present CausalGloss, an
English-prefix-to-ASL-gloss action policy for a Mandarin-speech cascade. It
combines lexical and semantic evidence, bounded delay, negation constraints,
burst WRITE actions, and incomplete-history recovery. On a frozen 500-example
How2Sign-derived benchmark with weak LLM-generated references, CausalGloss
improves the fixed Wait-$k$ frontier and nearly matches a three-seed
TransLLaMA-style learned adaptive baseline at $k=5$ in Token F1 (0.7047 versus
0.7067), with 54.4\% lower LAAL. Against CLERC Epee human-gloss references, it
achieves 0.649 Token F1 and Pareto-dominates three-seed TransLLaMA-style $k=5$
and fixed $k=5,7$, reducing LAAL by 58.2--69.9\%. Ablations separate timing
from retention. A synthetic Mandarin-speech cascade preserves a smaller
computation-aware latency gain, although outputs remain over-long. We also
construct ChineseSpeech-ASLGloss-v1, a 19,924-utterance, 36.54-h controlled
evaluation corpus.
\end{abstract}
\endgroup


\section{Introduction}
\label{sec:introduction}

Real-time spoken-to-signed translation requires target signs before an
utterance ends. Most simultaneous systems translate speech incrementally into
speech or text
\cite{ma-etal-2019-stacl,koshkin-etal-2024-transllama,ouyang-etal-2025-infinisst},
while simultaneous sign-language systems map sign video to spoken or written
text \cite{yin-etal-2021-simulslt,sun-etal-2024-adaptive}. Opposite-direction
systems generate signs from complete speech or text through gloss, pose, or
video \cite{ma-etal-2024-ms2sl,moryossef-etal-2023-open}. These systems do not
directly address when each ASL gloss can be committed from evolving speech.

English and ASL gloss differ in order, length, and lexical form, here termed
source--gloss divergence
\cite{moryossef-etal-2021-data,varanasi-etal-2024-linguistically}. Wait-$k$
uses a fixed positional schedule and cannot adapt each decision to distant
source--gloss evidence. Learned policies instead need causal alignment: the
earliest observed prefix that licenses a target gloss
\cite{ma-etal-2019-stacl,koshkin-etal-2024-transllama}. A contextual anchor,
or source position selected by multilingual contextual embeddings
\cite{jalili-sabet-etal-2020-simalign}, does not by itself guarantee semantic
sufficiency for late negation or repeated words. Teacher-forced trajectories
also omit inference-time error states
\cite{bengio-etal-2015-scheduled}.

How2Sign pairs ASL video with English, whereas CLERC Epee pairs English prompts
with human glosses \cite{duarte-etal-2021-how2sign,clerc-epee-2026}; neither
covers this setting directly. We study an explicitly scoped cascade: Mandarin
speech is translated incrementally into English, then an English-prefix policy
decides when to emit ASL glosses. This separates speech processing from action
control and tests its trade-off against fixed and learned adaptive policies.

Our contributions are threefold: (i) a simultaneous Mandarin speech-to-ASL
gloss cascade from streaming speech to online gloss actions; (ii) CausalGloss,
which combines lexical, semantic, delay, and negation evidence with burst
WRITE (consecutive glosses without another READ) and recovery from incomplete
histories; and (iii) ChineseSpeech-ASLGloss-v1, a 19,924-utterance corpus, plus
a locked evaluation with fixed and three-seed TransLLaMA-style learned curves,
paired intervals, and zero-shot human-gloss baseline comparisons.

\section{Related Work}
\label{sec:related_work}

\noindent\textbf{Simultaneous translation policies.}
Wait-$k$ fixes a prefix schedule through $k$ \cite{ma-etal-2019-stacl}.
Adaptive controllers decouple READ/WRITE decisions from Wait-$k$ schedules
\cite{zhao-etal-2023-adaptive}. TransLLaMA learns waiting from causal
alignments and uses $k$ as a minimum delay
\cite{koshkin-etal-2024-transllama}; InfiniSST handles unbounded speech with
multi-turn trajectories \cite{ouyang-etal-2025-infinisst}. We specialize
action control to sparse, reordered ASL-gloss targets.

\noindent\textbf{Signed-language translation.}
Existing simultaneous sign-language translation maps sign video to incremental
spoken or written language \cite{yin-etal-2021-simulslt,sun-etal-2024-adaptive}.
Speech/text-to-sign systems instead generate signs from complete input through
gloss, pose, or video \cite{ma-etal-2024-ms2sl,moryossef-etal-2023-open,
moryossef-2024-sign}, while linguistically informed text-to-gloss models remain
sentence-level \cite{varanasi-etal-2024-linguistically}. Gloss--text
divergence denotes differences in lexical choice, order, and length between
the representations; it motivates data augmentation and scaled
back-translation \cite{moryossef-etal-2021-data,ye-etal-2023-scaling}.
SimAlign provides source positions from multilingual contextual embeddings
\cite{jalili-sabet-etal-2020-simalign}; CausalGloss combines contextual
evidence with lexical occurrence, semantic sufficiency, bounded delay, and
negation constraints. Recovery states address the
teacher-forced/free-running mismatch \cite{bengio-etal-2015-scheduled}. LAAL
avoids rewarding over-generation with artificially low latency
\cite{papi-etal-2022-generation}.

\section{Proposed Method}
\label{sec:method}

Let $\mathbf{x}=(x_1,\ldots,x_m)$ be an English source and
$\mathbf{y}=(y_1,\ldots,y_n)$ its ASL-gloss target. Causal alignment assigns
each gloss its earliest licensing prefix. In the full cascade, streaming
Mandarin is first translated into incremental English, on which CausalGloss
operates.

\subsection{Causal action policy}

At step $t$, $i_t$ counts observed source tokens and $j_t$ indexes the next
gloss. The state contains $\mathbf{x}_{\leq i_t}$, committed history
$\hat{\mathbf{y}}_{<j_t}$, and a completion flag. Action $a_t$ is
\texttt{<WAIT>}, \texttt{WRITE($g$)}, or \texttt{<END>}, with
\begin{equation}
(i_t,j_t)\xrightarrow{a_t}
\begin{cases}
(\min(i_t+1,m),j_t), & a_t=\mathrm{WAIT},\\
(i_t,j_t+1), & a_t=\mathrm{WRITE}(g),\\
\text{terminate}, & a_t=\mathrm{END}.
\end{cases}
\label{eq:transitions}
\end{equation}
Here $g$ is the next gloss. WRITE leaves $i_t$ unchanged, so consecutive
WRITEs at one prefix form a burst. At source completion, the policy emits any
remaining glosses before \texttt{<END>}.

\subsection{Multi-evidence causal alignment}

For target gloss $y_j$, we obtain a lexical anchor $a_j$ from normalized exact
and Porter-stem matches; Porter stemming removes suffixes so forms such as
\emph{walk} and \emph{walking} can match. Anchor $a_j$ is the earliest
compatible source position for gloss $j$. Occurrence-aware matching prevents
repeated source words from reusing an earlier occurrence, while a multi-piece
gloss requires its pieces in local source order. Full pairs are used only to
construct training alignments; inference sees prefixes. An unmatched gloss is
placed monotonically between its nearest matched neighbors, or at the nearest
boundary, without a lexical-evidence flag.

Lexical presence alone does not imply semantic sufficiency. A frozen
English-to-gloss generator scores every source prefix. If
$(b_{j1},\ldots,b_{jK_j})$ are the target subtokens of $y_j$, we use
\begin{equation}
s_j(i)=\frac{1}{K_j}\sum_{r=1}^{K_j}
\log p_{\phi}(b_{jr}\mid\mathbf{x}_{\leq i},\mathbf{y}_{<j},b_{j,<r}).
\label{eq:prefix_score}
\end{equation}
This measures prefix support relative to full-source score $s_j(m)$. A prefix
is stable when its score remains above
$\max\{s_j(m)-\delta_s,\log p_{\min}\}$ for $w$ consecutive prefixes,
where $w$ controls the required persistence. Strong exact or stem anchors emit at $a_j$ when
the full-to-prefix gap is below $\delta_{\mathrm{lex}}$; otherwise we use the
earliest stable prefix. If it lags over $D$ tokens behind $a_j$, bounded-delay
rescue permits $a_j+D$ only under the looser gap $\delta_{\mathrm{res}}$.
A negated gloss cannot precede its observed negation. Positions are made
nondecreasing; ties create burst WRITEs in Eq.~\ref{eq:transitions}.

\subsection{Recovery from free-running histories}

Teacher forcing presents correct previous actions and glosses, whereas
free-running inference uses model history. An unnecessary WAIT can leave
glosses overdue, a state absent from training. If two glosses are licensed but
only the first is committed, recovery supervision asks the model to WRITE the
second before reading again.

We add four valid off-policy states: complete source with (i) empty, (ii)
midpoint, or (iii) one-gloss-lagged history, and (iv) partial source with
multiple glosses due. They teach catch-up WRITEs without changing the target.
In our comparison, \emph{action-only} uses aligned states,
\emph{guard-only} also forbids WAIT after source completion, and
\emph{recovery-trained} learns from all four added states.

The action model uses a teacher-forced language-model loss,
\begin{equation}
\mathcal{L}_{\mathrm{act}}=-\sum_t
\log\pi_{\theta}(a_t^*\mid\mathbf{x}_{\leq i_t},
\mathbf{y}_{<j_t},c_t),
\label{eq:action_loss}
\end{equation}
where $c_t$ marks source completion. Instruction, source, and history tokens
are context but masked from loss; only the correct action and, for WRITE, its
gloss are targets. Free-running evaluation uses model outputs as history.

\section{Experimental Setup}
\label{sec:experiments}

\noindent\textbf{Data and protocol.}
We use manually realigned English transcriptions from How2Sign
\cite{duarte-etal-2021-how2sign} and ASL-gloss targets deterministically
generated by a frozen Qwen2.5-7B adapter. These targets provide weak
supervision and are not expert-gold ASL annotations. Source video IDs are
disjoint across splits. The action-training set contains 18,632 parent
utterances. A separate, fixed 500-parent pool is split equally into development
and diagnostic subsets; an independent 500-parent test set remains locked.
Only development selects checkpoints. The diagnostic subset is not the test
set and is used only for component and recovery ablations.

\noindent\textbf{Model and optimization.}
The action policy is Qwen2.5-7B-Instruct \cite{yang-etal-2024-qwen25},
initialized from the development-selected pre-recovery adapter. We apply
low-rank adaptation (LoRA) \cite{hu-etal-2022-lora} to query and value
projections with rank 16, $\alpha=32$, and dropout 0.05. The 439,103 actions
comprise 367,267 balanced aligned states and 71,836 recovery states. We train
with AdamW at $10^{-5}$ learning rate, 0.01 weight decay, 3\% warm-up, unit
gradient clipping, effective batch size 32, and bfloat16 precision. We evaluate
every 1,000 updates after step 4,000 and stop at 6,000. S1, S2, and S3 denote
seeds 20260829, 20260904, and 20260905. They share data hashes and
configurations; S1/S2 use two RTX 5090 GPUs and S3 two A800 GPUs. The frozen
selection order is Token F1, sequence similarity, then LAAL.

The frozen Qwen2.5-7B scorer in Eq.~\ref{eq:prefix_score} generated the weak
targets after one epoch on 27,079 pairs. Sharing it across target construction
and alignment creates dependence that external human-gloss evaluation only
partly addresses.
We set $\delta_s=0.75$, $\delta_{\mathrm{lex}}=0.25$, $p_{\min}=0.01$, and
$\delta_{\mathrm{res}}=1.0$, with $w=1$ and $D=3$.

\noindent\textbf{Fixed Wait-$k$ baselines.}
We train schedule-specific policies with
$g(t)=\min(k+t-1,m)$ \cite{ma-etal-2019-stacl}, for $k\in\{1,3,5,7\}$.
WRITE preserves the prefix; \texttt{<WAIT>} reads one token. All use the same
Qwen2.5-7B backbone, upstream English--gloss data, LoRA, effective batch size
32, splits, and greedy free-running decoding as CausalGloss. Fixed policies
train one epoch at $3\times10^{-5}$; CausalGloss recovery runs 6,000 updates at
$10^{-5}$. These are complete training recipes, not isolated trajectory
comparisons. All schedules were predeclared; $k=5,7$ use three seeds.

\noindent\textbf{Learned adaptive baseline.}
Following TransLLaMA \cite{koshkin-etal-2024-transllama}, we independently
derive SimAlign causal trajectories in which \texttt{<WAIT>} reads and a gloss
writes. The baseline shares the Qwen2.5 backbone, pseudo-ASL data, tokenizer,
upstream adapter, LoRA, and splits. A one-seed validation-loss screen selects
the learning rate from $\{10^{-5},3\!\times\!10^{-5},6\!\times\!10^{-5}\}$;
three final seeds train three epochs. We predeclare the minimum initial-read
$k=1,\ldots,8$ and decode the locked 500 free-running without gold histories.
This is a task/backbone adaptation, not an exact reproduction.

\noindent\textbf{Metrics and statistics.}
Token F1 uses multiset token overlap; sequence similarity is the order-sensitive
token-sequence \texttt{SequenceMatcher} ratio. A rollout is complete only if it
emits \texttt{<END>} after consuming the full source. LAAL
\cite{papi-etal-2022-generation} is measured in source tokens, or seconds for
speech. Post-hoc BLEU-4 uses untokenized, unsmoothed, single-reference scoring.
Locked-500 fixed-policy differences use 10,000 paired-parent bootstrap
resamples. Learned-baseline and CLERC three-seed comparisons use 20,000
hierarchical draws over seeds and paired samples; CausalGloss and fixed-policy
seeds are independently resampled. Fixed $k=1,3$ Epee intervals resample
prompts only. The predeclared retention margin is $-0.03$ Token F1.

\noindent\textbf{External and cascade evaluation.}
With decisions frozen, we evaluate 401 CLERC Epee v0.3 prompts against 1,200
gloss sequences from six signers \cite{clerc-epee-2026}. CausalGloss,
TransLLaMA-style $k=1,\ldots,8$ (confirmatory $k=5$; others descriptive), and
fixed $k\in\{1,3,5,7\}$ share normalization, greedy decoding, budgets, and
metrics; references serve only for scoring. All but fixed $k=1,3$ use three
seeds. Before best-reference scoring, we remove \texttt{QUESTION} tags,
map numbered variants to base glosses, and collapse exact repetitions. We also
construct
ChineseSpeech-ASLGloss-v1, containing 19,924 synthetic Mandarin utterances
(36.54 h) from four voices with fixed 18,924/500/500 splits. Its locked test is
processed in 1.2-s chunks by SeamlessStreaming 2.5B
\cite{barrault-etal-2023-seamless} to produce incremental English before action
decoding. Computation-aware LAAL includes measured serial inference time; we
also report end-to-end real-time factor (RTF). The corpus is synthetic and its
gloss targets come from the same frozen generator.

\section{Results}
\label{sec:results}

\begin{figure*}[t]
  \centering
  \includegraphics[width=\textwidth]{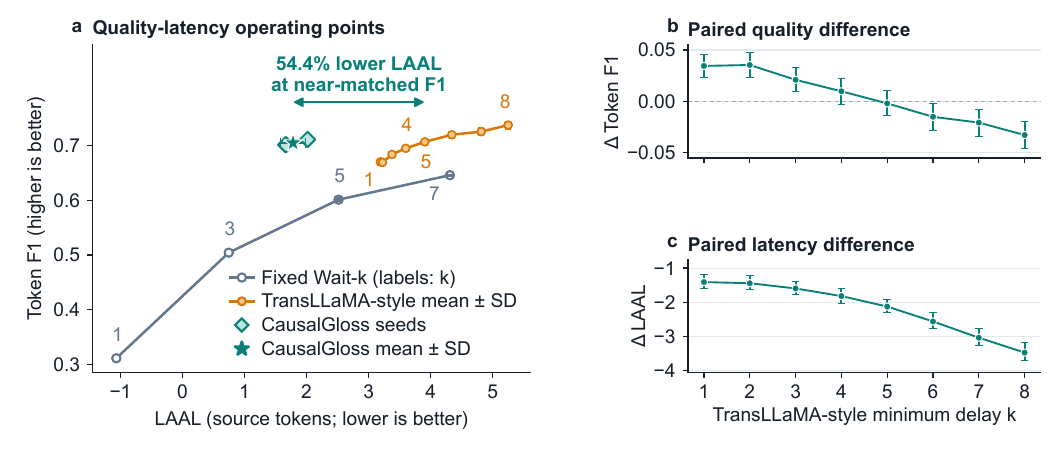}
  \caption{Locked-500 results. (a) Fixed Wait-$k$, TransLLaMA-style learned
  $k=1,\ldots,8$ (three-seed means $\pm$ SD), and CausalGloss seeds and mean;
  labels denote $k$. (b,c) CausalGloss minus learned baseline, with hierarchical
  seed-and-parent bootstrap 95\% CIs.}
  \label{fig:quality_latency}
\end{figure*}

\subsection{Quality--latency trade-off}

\begin{table}[t]
  \caption{Locked-500 comparison: three-seed means and sample SDs;
  BLEU-4 is post-hoc.}
  \label{tab:main_locked500}
  \centering
  \fontsize{9.1pt}{11pt}\selectfont
  \setlength{\tabcolsep}{1.5pt}
  \begin{tabular*}{\columnwidth}{@{\extracolsep{\fill}}lrrr@{}}
    \toprule
    & \multicolumn{3}{c}{System} \\
    & Wait-$k=7$ & \shortstack{TransLLaMA-\\style $k=5$} & CausalGloss \\
    \midrule
    F1 $\uparrow$ & $.6457\!\pm\!.0010$ & $.7067\!\pm\!.0056$ & $.7047\!\pm\!.0055$ \\
    Sim. $\uparrow$ & $.6383\!\pm\!.0008$ & $.6955\!\pm\!.0051$ & $.6957\!\pm\!.0060$ \\
    BLEU-4 $\uparrow$ & $33.90\!\pm\!.19$ & $28.35\!\pm\!.33$ & $37.31\!\pm\!2.88$ \\
    LAAL $\downarrow$ & $4.3099\!\pm\!.0528$ & $3.9061\!\pm\!.0458$ & $1.7823\!\pm\!.2031$ \\
    \bottomrule
  \end{tabular*}
\end{table}

Fixed Wait-$k$ showed the expected quality--latency trade-off: Token F1 rose
from 0.3111 to 0.6457 as LAAL increased from $-1.0660$ to 4.3099 source tokens
(Table~\ref{tab:main_locked500}; Fig.~\ref{fig:quality_latency}a). CausalGloss
achieved $0.7047\pm0.0055$ Token F1 and $0.6957\pm0.0060$ sequence similarity
at $1.7823\pm0.2031$ LAAL across three seeds. Every seed had higher F1 and
lower LAAL than the two highest-quality fixed points ($k=5,7$), thereby
improving the empirical frontier. Relative to the three-seed $k=7$ mean,
Token F1 increased by 0.0590 and LAAL decreased by 2.5276 source tokens, or
58.6\%.

Fixed-$k=7$ aggregate 95\% CIs excluded zero for F1 ($[.039,.079]$) and
LAAL ($[-2.830,-2.197]$); all seed-wise CIs did likewise.

Against the TransLLaMA-style baseline, CausalGloss nearly matched $k=5$ mean
F1 (0.7047 versus 0.7067) while reducing LAAL by 2.1235 tokens, or 54.4\%.
The hierarchical 95\% CI included zero for F1 ($[-.0139,.0102]$) but excluded
zero for LAAL ($[-2.3010,-1.9237]$). CausalGloss mean Pareto-dominated learned $k=1$--4; learned $k=6$--8 gained
0.0150--0.0328 F1 at 2.5585--3.4677 more LAAL. All learned settings completed
98.5--98.9\%, with mean invalid-action rates below 0.53\%.

\subsection{External and cascade evaluation}

On CLERC Epee, CausalGloss reached 0.6490 F1 at 1.5388 LAAL and
Pareto-dominated learned $k=1,\ldots,8$. At confirmatory $k=5$, F1 rose
0.1001 (95\% CI $[.082,.118]$), LAAL fell 58.2\%, and completion was 100\%
versus 87.0\%. It dominated fixed $k=3,5,7$, but not $k=1$; versus
predeclared three-seed $k=7$, F1 rose 0.0404 (95\% CI $[.016,.066]$) and
LAAL fell 69.9\%. As a post-hoc robustness check, on prompts for which the TransLLaMA-style
$k=5$ baseline completed, CausalGloss still achieved higher F1 (0.6196 vs.
0.5839; $\Delta=0.0357$, 95\% CI $[.021,.051]$; $n=344$--$353$ per seed).

On locked synthetic Mandarin speech, predeclared S1 retained F1 relative to
the earlier lexical-only controller (0.3946 versus 0.3941; $\Delta=+0.0004$,
95\% CI $[-0.0084,0.0095]$) while reducing LAAL by 0.152\,s (95\% CI
$[-0.220,-0.092]$). Accounting for processing, the gain was 0.123\,s
(95\% CI $[-.193,-.063]$), modest relative to 1.2-s chunks.

Across seeds, F1 was $0.3944\pm0.0003$, raw LAAL was $2.415\pm0.042$\,s, and
completion was 99.8\%; RTF was 0.388 for the lexical-only controller and 0.380
for CausalGloss. Every seed failed the output-length criterion, confirming
over-generation as the system-level failure.

\subsection{Component ablation}

Ablating semantic evidence or burst WRITE increased LAAL by 0.960 and 1.624
tokens without significant F1 changes (Table~\ref{tab:component_ablation}).
Ablating rescue, negation, occurrence matching, or recovery reduced F1 by
0.062, 0.072, 0.073, and 0.075; all but occurrence matching also increased
LAAL by 0.326, 0.267, and 0.306 tokens. Bold effects excluded zero, and all
variants completed 100\%. Recovery remained 0.038 F1 above guard-only,
supporting learning beyond the completion rule.

\begin{table}[t]
  \caption{Component ablation on the frozen diagnostic set. Values are
full-minus-ablated (action-only for recovery); bold: 95\% CI excludes zero.
Full F1/LAAL = 0.7131/1.5146.}
  \label{tab:component_ablation}
  \centering
  \fontsize{9.1pt}{11pt}\selectfont
  \setlength{\tabcolsep}{2.0pt}
  \renewcommand{\arraystretch}{1.0}
  \begin{tabular*}{\columnwidth}{@{\extracolsep{\fill}}lcc@{}}
    \toprule
    Removed component & $\Delta$F1 $\uparrow$ & $\Delta$LAAL $\downarrow$ \\
    \midrule
    Semantic evidence & +0.008 & \textbf{-0.960} \\
    Bounded-delay rescue & \textbf{+0.062} & \textbf{-0.326} \\
    Negation constraint & \textbf{+0.072} & \textbf{-0.267} \\
    Occurrence-aware matching & \textbf{+0.073} & -0.252 \\
    Burst WRITE & +0.003 & \textbf{-1.624} \\
    Recovery states & \textbf{+0.075} & \textbf{-0.306} \\
    \bottomrule
  \end{tabular*}
\end{table}

\noindent\textbf{Why the frontier moves.}
The fixed curve establishes the schedule trade-off; the learned curve directly
tests adaptive control. Near learned $k=5$ quality, CausalGloss's lower latency
and high completion rule out failed rollouts as the explanation. It instead
gates WRITE on lexical, semantic, and delay evidence. Ablations separate release
timing (semantic evidence, burst WRITE) from early-commitment coverage (rescue,
negation, occurrence matching, recovery); guard-only controls exclude trivial
truncation.

\noindent\textbf{From controller to cascade.}
The Mandarin cascade retains our earlier lexical-only controller as an internal
control, not an external baseline. S1's 0.152-s raw gain remains 0.123\,s with
processing, while similar RTFs and 1.2-s chunks bound its magnitude. Epee
measures automatic overlap with human annotations, not expert preference or
non-manual marking. The learned baseline is task-adapted, not an exact
TransLLaMA reproduction; comparisons cover complete training recipes rather
than trajectory supervision alone. Expert annotation and length calibration
remain necessary.

\section{Conclusion}
\label{sec:conclusion}

Across three seeds, CausalGloss cuts LAAL by 58.6\% versus fixed $k=7$ and
54.4\% at near-matched F1 versus learned $k=5$. On CLERC Epee, both metrics
improve, supporting zero-shot transfer to human-gloss references. Ablations
separate timing from retention; speech yields a 0.123\,s latency gain. Expert ASL validation and over-generation remain open.

\clearpage
\section{Compliance with Ethical Standards}
This study involved no new collection of data from human participants. It used
previously released How2Sign and CLERC Epee data in accordance with their
respective licenses and terms of use, together with synthetically generated
speech. No additional ethical approval was required for this secondary
computational study.

{\noindent\textbf{Acknowledgment.}
The authors would like to thank the support from Project W2531054 of the
National Natural Science Foundation of China and the Program for Guangdong
Introducing Innovative and Entrepreneurial Teams.
The authors declare no conflicts of interest.\par}

\bibliographystyle{IEEEbib}
\bibliography{refs}

\end{document}